\documentclass[runningheads]{llncs}
\usepackage{colortbl} 
\usepackage{xcolor}
\usepackage{array}
\usepackage{graphicx}
\usepackage{booktabs} 
\usepackage{multirow} 
\usepackage{bbm} 
\usepackage{amssymb}
\usepackage{wrapfig} 
\usepackage{array}
\usepackage{algorithm}
\usepackage{marvosym}
\usepackage[marginal]{footmisc}

\usepackage{bm}
\usepackage{amsmath}
\usepackage{hyperref}
\usepackage{pifont} 
\usepackage{mathtools} 
\usepackage{color}

\begin{document}
\setlength{\textfloatsep}{10pt}
%
\title{VS-Assistant: Versatile Surgery Assistant on the Demand of Surgeons}
\titlerunning{Versatile Surgery Assistant on the Demand of Surgeons}
%
\author{Zhen Chen\inst{1}{*}, Xingjian Luo\inst{1}{*}, Jinlin Wu\inst{1}, Danny T.M. Chan\inst{2}, Zhen Lei\inst{1}, Jinqiao Wang\inst{1}, Sebastien Ourselin\inst{3}, Hongbin Liu\inst{1}}
%
\authorrunning{Z. Chen et al.}
%
\institute{Centre for Artificial Intelligence and Robotics (CAIR), HKISI-CAS \and
    Dept. of Surgery, The Chinese University of Hong Kong \and King's College London} 
%
\maketitle              
%
\begin{abstract}
{The surgical intervention is crucial to patient healthcare, and many studies have developed advanced algorithms to provide understanding and decision-making assistance for surgeons. Despite great progress, these algorithms are developed for a single specific task and scenario, and in practice require the manual combination of different functions, thus limiting the applicability. Thus, an intelligent and versatile surgical assistant is expected to accurately understand the surgeon's intentions and accordingly conduct the specific tasks to support the surgical process. In this work, by leveraging advanced multimodal large language models (MLLMs), we propose a Versatile Surgery Assistant (VS-Assistant) that can accurately understand the surgeon's intention and complete a series of surgical understanding tasks, \textit{e.g.}, surgical scene analysis, surgical instrument detection, and segmentation on demand. Specifically, to achieve superior surgical multimodal understanding, we devise a mixture of projectors (MOP) module to align the surgical MLLM in VS-Assistant to balance the natural and surgical knowledge. Moreover, we devise a surgical Function-Calling Tuning strategy to enable the VS-Assistant to understand surgical intentions, and thus make a series of surgical function calls on demand to meet the needs of the surgeons. Extensive experiments on neurosurgery data confirm that our VS-Assistant can understand the surgeon's intention more accurately than the existing MLLM, resulting in overwhelming performance in textual analysis and visual tasks. Source code and models will be made public.}


\keywords{Large Language Models \and Multimodal LLM \and Surgery understanding \and Neurosurgery.}
\end{abstract}
\section{Introduction} 

The computer-assisted surgery stands at the forefront of enhancing interventional healthcare, thereby facilitating patient safety and outcomes \cite{maier2022surgical,lam2022machine,ban2023concept}. In particular, deep learning algorithms extensively explored to support the surgical process from many aspects, \textit{e.g.}, monitoring surgical procedures \cite{jin2021temporal}, optimizing surgeon scheduling \cite{abdalkareem2021healthcare}, enhancing team coordination \cite{kennedy2020computer}, and advancing the training of junior surgeons  \cite{lee2021artificial}.

Current research has concentrated on harnessing surgical endoscopic videos and images, aiming to provide precise understanding and decision-making assistance for surgeons. Notably, Das \textit{et al.} \cite{das2023multi} designed a dual-head network to segment critical anatomical structures and their regional centroids in endoscopic pituitary surgery. Chen \textit{et al.} \cite{chen2023surgical} interpreted the surgical actions in natural language descriptions by bridging the modality gap with surgical concepts. Seenivasan \textit{et al.} \cite{seenivasan2023surgicalgpt} proposed a GPT-2 based language-vision model that excels in visual question answering within the surgical context, and Bai \textit{et al.} \cite{bai2023cat} further localized the targeted surgical instruments by responding to surgical queries. However, these advancements are hindered by a common limitation: \textit{these surgical algorithms are designed to handle a single task, and cannot be performed when needed according to the surgeon's intention}. To address this problem, a unified surgical system is expected to enable seamless communication and query handling, thereby accommodating diverse functions on the demand of surgeons. Therefore, such a unified surgical system would streamline AI applications in surgery, fostering a dynamic and responsive environment where technology and human expertise synergistically enhance the surgical experience and patient healthcare.




As an emergent technology with substantial attention, Large Language Models (LLMs) \cite{gpt35,llama2} have demonstrated remarkable capabilities in comprehension, reasoning, and planning, and are capable of parsing the intention of language queries. On this basis, Multimodal Large Language Models (MLLMs) \cite{llava,qwen,gpt4} further provide visual perception to handle images and videos, which are the foundational capabilities for the unified surgical system. Despite great progress, the current LLMs and MLLMs studies cannot address the complex demands of surgical operating rooms with three primary shortcomings. Firstly, existing LLMs lack specialized knowledge in surgery and cannot understand the context related to surgical scenarios. Secondly, there is a lack of visual comprehension of surgical scenes, and a more adequate multi-modal alignment is needed considering the variability of surgical visual contents. Lastly, they lack the capability to invoke different surgical algorithms according to the surgeon's intention, which is a critical functionality for responsive and adaptive surgical assistance.

To achieve such a goal, we propose a Versatile Surgery Assistant, named VS-Assistant, by leveraging the capability of multimodal large language models (MLLMs). The VS-Assistant can accurately understand the surgeon's intention and complete a series of surgical understanding tasks, \textit{e.g.}, surgical scene analysis, surgical instrument detection, and segmentation on demand. Specifically, we first devise the surgical LLM tuning strategy to produce the tailored LLM with surgical professional knowledge. To achieve superior surgical multimodal understanding, we devise a mixture of projectors (MOP) module with the dynamic routing strategy to align the surgical MLLM in VS-Assistant to balance the natural and surgical knowledge. Moreover, we devise a surgical Function-Calling Tuning strategy to enable the VS-Assistant to understand surgical intentions in three steps of \textit{Thinking}, \textit{Calling}, and \textit{Replying}, and thus make a series of surgical function calls on demand to meet the needs of the surgeons. Extensive experiments on neurosurgery data confirm that our VS-Assistant can understand the surgeon's intention more accurately than the existing MLLM, resulting in overwhelming performance in textual analysis and visual tasks.

\section{Versatile Surgery Assistant}
\begin{figure}[t]
  \centering
  \centerline{\includegraphics[width=0.99\textwidth]{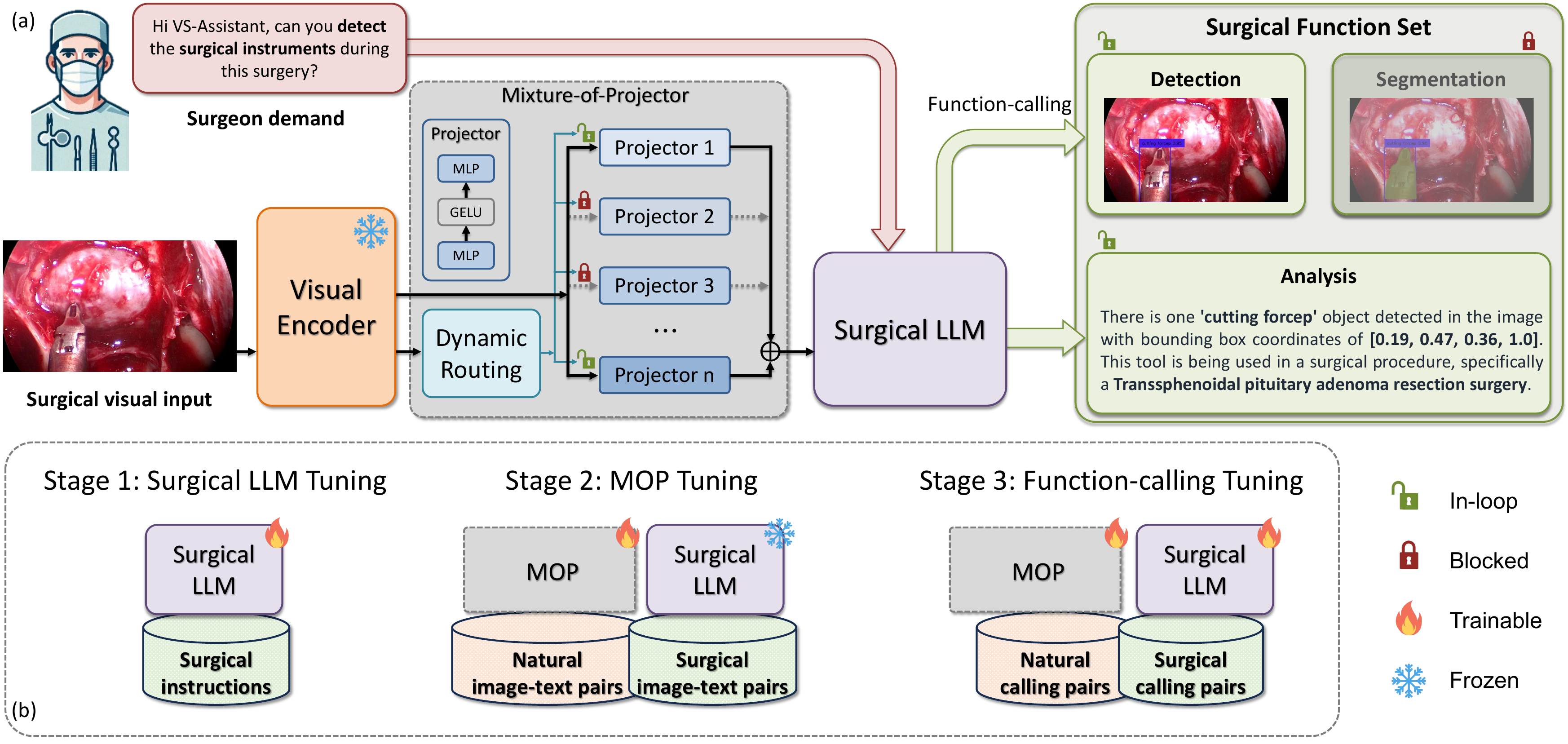}}
\caption{(a) The VS-Assistant framework consists of the visual encoder, the mixture-of-projector (MOP), the surgical LLM and the surgical function set. The VS-Assistant is capable to accomplish diverse surgical tasks on the demand of surgeons. (b) The training pipeline of VS-Assistant, including surgical LLM tuning, MOP tuning, and function-calling tuning.}
\label{model_arch}
\end{figure}

The proposed Versatile Surgery Assistant (VS-Assistant) in Fig.~\ref{model_arch} (a) is capable of comprehending multimodal surgical queries on the demand of surgeons. Our VS-Assistant consists of four modules, including the visual encoder, the surgical LLM, the mixture-of-projectors (MOP), and the surgical function set. Given the surgical visual input, the MOP coordinates the visual encoder and surgical LLM, and the surgical LLM parses the surgeon's demands based on the visual context, thereby calling corresponding surgical functions on the demand of surgeons and enhancing the performance of the final reply.


\subsection{Surgical LLM with Professional Knowledge}\label{llm}
In our VS-Assistant, we devise the surgical LLM to understand the surgical context with professional knowledge. Specifically, we first collect a large-scale corpus of surgical textbooks and construct the surgical instruction dataset of $30,000$ question-answer pairs. Then, we fine-tune the pre-trained LLM \cite{vicuna} to enhance the capability of answering surgery-related questions, as follows:
\begin{equation}
    L_{\rm SurgLLM} = -\sum \limits_{t=1}^{T} \text{log }p(y_t|y_{<t})
\end{equation}
where $y_{<t}$ is the previous predicted tokens before current token $t$, and $T$ is the token number of prediction. In this way, the surgical LLM is expanded with the semantic scope of surgical professional knowledge, thereby serving as the basis for the VS-Assistant. 



\subsection{Surgical Multimodal Alignment with Mixture of Projectors}\label{projector}
To obtain the multimodal capability, the current MLLMs align the visual features from the visual encoder and LLMs with different designs of the projector, including the linear layer \cite{llava}, Q-Former \cite{blip2} and several cross-attention layers \cite{flamingo} between the hidden layers of LLM and the visual encoder. Although effective in natural scenes, these methods adopted a single static projector, which limits the capabilities of the entire MLLM on more delicate surgical scenes. To empower the surgical LLM with multimodal understanding, we propose a Mixture of Projectors (MOP) module to 
dynamically align visual features from the visual encoder with the semantic space of the surgical LLM.

Given surgical visual input, the visual encoder transforms it into ${I} \in \mathbb{R}^{L_I \times C_I}$, and concurrently the surgeon query is tokenized into $Q \in \mathbb{R}^{L_Q \times C_Q}$ as the input of surgical LLM, where $L_I$ and $L_Q$ are the number of tokens,  and $C_I$ and $C_Q$ are the embedding dimensions. To map the visual embeddings $I$ into the space of surgical LLM, the MOP consists of $N$ parallel projectors to process the visual embeddings at the same time, and each projector is a two-layer multi-layer perceptron (MLP) with activation function as follows:
\begin{equation}
    {P}({I}) = \mathrm{MLP}(\varphi(\mathrm{MLP}({I}))), 
\end{equation}
where $\mathrm{MLP}$ is a fully-connected layer and $\varphi$ is the GeLU activation. Meanwhile, to adaptively coordinate visual and language modalities based on the input, the dynamic routing module in MOP determines the optimal $K$ projectors and calculates the dynamic routing weights $D$, as follows:
\begin{equation}
    {D}({I}) = \text{Softmax}(\text{topK}(f_{\rm con}({I}+\mathcal{\xi}))),
\end{equation}
where $f_{\rm con}$ is a MLP to identify the contribution of projectors on the input, and $\mathcal{\xi}\sim \mathcal{N}(0,\sigma^2)$ represents Gaussian noise with the standard deviation $\sigma$ to augment the MOP tuning. Finally, the output visual token ${I}_{\rm MOP} \in \mathbb{R}^{L_I \times C_Q}$ of the MOP is calculated by aggregating the outputs of $K$ selected projectors with the dynamic routing weights, as follows:
\begin{equation}
     I_{\rm MOP} = \sum_{k=1}^{K}  D_k(I) \cdot  {P}_k(I),
\end{equation}
where $P_k$ is selected by the topK operation, and $D_k$ is the corresponding dynamic routing weight. In this way, each projector in the MOP is specialized to handle distinct aspects of the surgical scenarios, and the dynamic routing module determines the most appropriate projector combinations for each input.

\subsection{Surgical Function-Calling Tuning}\label{ft}




To enable the VS-Assistant to accomplish the requirement of surgeons, we propose the surgical Function-Calling Tuning strategy to autonomously determine the necessity of invoking surgical functions. Specifically, we maintain a surgical function set composed of diverse algorithms, \textit{e.g.}, surgical scene analysis, instrument detection, and segmentation, and formulate the function calling as generating the name of executable Application Programming Interface (API) to receive the function output. Instead of directly fine-tuning the MLLM with function conversation samples, we tailor the function-calling conversation based on the Chain-of-Thought (CoT) \cite{cot} to guide the VS-Assistant accurately understand the surgeon's intention and conduct the required functions.

In the function calling of the VS-Assistant, we structure the conversation into three pivotal components, including the \textit{Thinking}, \textit{Calling}, and \textit{Replying}. As elaborated in Fig.~\ref{fig:function-calling}, the \textit{Thinking} provides the VS-Assistant with the function-calling deliberation to support surgical decision-making. On the basis of \textit{Thinking}, the \textit{Calling} enables the VS-Assistant to directly execute surgical functions through the API name, without the need for parsing the entire textual output. The \textit{Replying} is the textual output for surgeons and provides the supervision for fine-tuning the VS-Assistant. In the function-calling tuning, the surgical LLM is delivered with multimodal input $X$ and generates the predictions of these three components, as $\mathcal{T}, \mathcal{C}, \mathcal{R}  = \text{SurgLLM}(X)$, where $\mathcal{T}$, $\mathcal{C}$ and $\mathcal{R}$ are the \textit{Thinking}, \textit{Calling}, and \textit{Replying}, respectively.  


\begin{figure}[t]
\centering 
\fbox{%
  \parbox{0.965\textwidth}{%
    \vspace{3pt} 
    \textbf{\textit{Thinking}}: Internal deliberations on whether to call a specific function or not. \\
    {\footnotesize \textbf{Example}: The utilization of a detection model to ascertain the presence of the target object could be highly beneficial.} \\


    \textbf{\textit{Calling}}: Specific API name and its associated parameters of executable functions.\\
    {\footnotesize\textbf{Example}: \{API name: Detection Function\}; \{API parameters: Navigation Probe\}.} \\
    
    \textbf{\textit{Replying}}: Textual output presented to surgeons ultimately. \\
    \textbf{Example}: The navigation probe is detected in the image, with bounding box coordinates of [0.18, 0.41, 0.45, 0.99]. The navigation probe is a medical instrument used during surgeries to sense and navigate through tissues. In the context of this image, the probe is being used during a Transsphenoidal pituitary adenoma resection surgery, and helps the surgeon to accurately locate the tumor ...
    \vspace{3pt} 
  }%
}
\caption{Components of CoT-based function-calling conversations.} 
\label{fig:function-calling} 
\end{figure}

\begin{figure}[t]
  \centering
  \centerline{\includegraphics[width=0.99\textwidth]{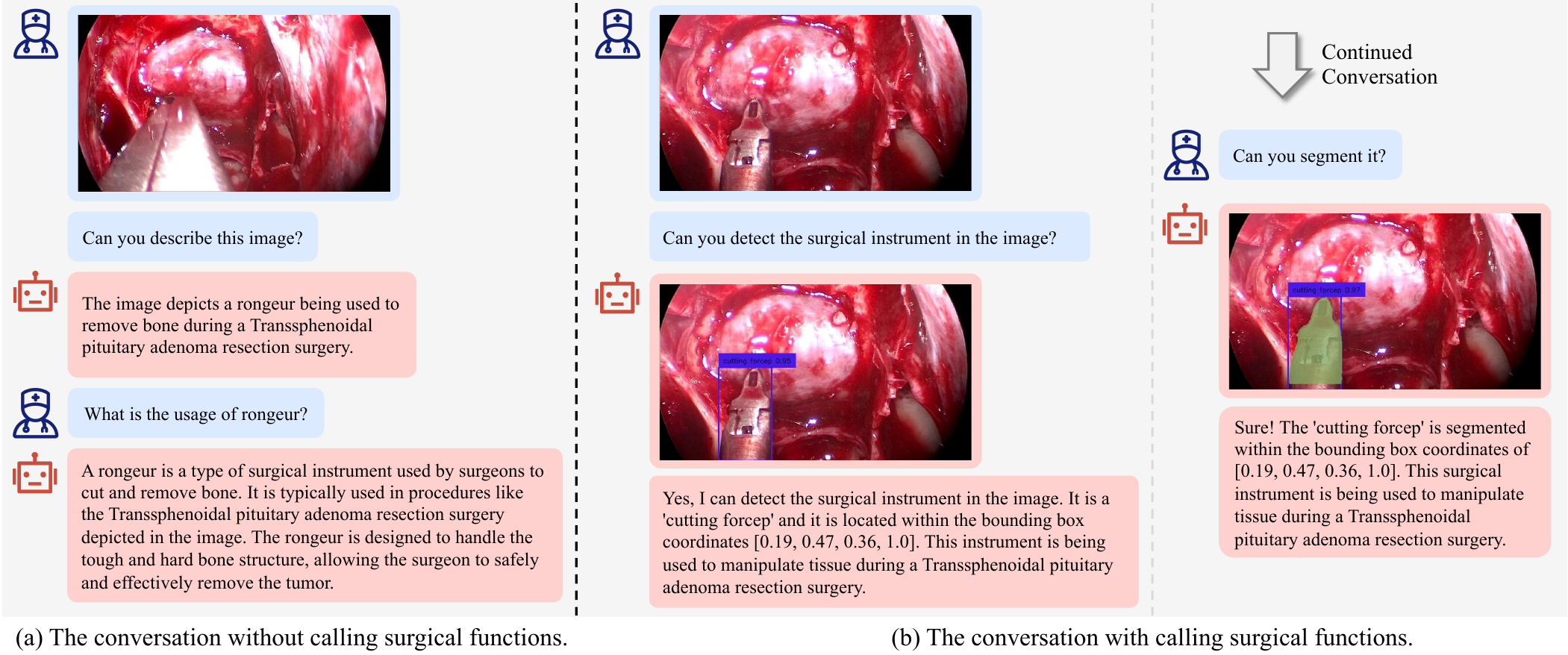}}
\caption{The conversation records of the VS-Assistant, including (a) without calling surgical functions and (b) calling surgical functions of detection and segmentation.}
\label{model_conversation}
\end{figure}

In our VS-Assistant, a single query from surgeons may elicit a one- or two-round conversation. If the VS-Assistant justifies it unnecessary to call functions (\textit{i.e.}, the output $\mathcal{C}$ is None), the VS-Assistant relies on its own multimodal capability and directly presents the predicted \textit{Replying} to surgeons, as Eq. \eqref{eq_output}.
In contrast, if the query exceeds the competence of MLLM, the VS-Assistant executes an appropriate function and utilizes the function results to infer again, thereby providing an enhanced \textit{Replying} as the textual output of VS-Assistant, as follows:    
\begin{equation}\label{eq_output}
    {\rm Textual \,\, Output} = 
    \begin{cases}   
    \mathcal{R}, & \text{if } \mathcal{C} \text{ is None} \\
    \mathcal{R} \in \text{SurgLLM}([X,\text{Execute}(\mathcal{C})]), & \text{otherwise}
    \end{cases}
\end{equation}
As elaborated in Fig.~\ref{model_conversation} (a), the VS-Assistant can analyze surgical visual input relying on its own MLLM capability and perform multi-round conversation. When surgical functions need to be called in Fig.~\ref{model_conversation} (b), the VS-Assistant can accurately understand the surgeon's intention and generate textual analysis based on the visual context returned by functions. In this way, our surgical function-calling tuning based on structured conversations endows the VS-Assistant with explicit reasoning, specifically, learning what, how, and when to call surgical functions.

\subsection{Optimization}\label{ft}
As illustrated in Fig.\ref{model_arch} (b), we train the VS-Assistant in three stages. In the first stage, we perform the surgical LLM tuning to adapt the LLM with surgical professional knowledge. Then, we freeze the surgical LLM and train the MOP from scratch using the image-text pairs of both natural and surgical scenarios, and perform the multimodal alignment to obtain a surgical MLLM. Finally, we perform the function-calling tuning with natural and surgical conversation dataset to empower the VS-Assistant with the capability of conducting surgical functions on the demand of surgeons.


\section{Experiment}
\subsection{Dataset and Implementation Details}

\noindent\textbf{Surgical LLM Tuning Dataset}.
We collect over $3,000$ surgical textbooks, and construct 30K surgical-related instructions (\textit{i.e.}, question-answer pairs) based on the parsed textbook contents, following the Stanford Alpaca protocol \cite{alpaca}. 

\noindent\textbf{MOP Tuning Dataset}.
We collect both natural and surgical image-text pairs to introduce multimodal knowledge in surgery while maintaining the capabilities of MLLM. 
We conduct the screening of these samples to ensure a relatively balanced proportion between the surgical data and the natural data. The total dataset contains a mixture of 595K natural datasets from LLaVA \cite{llava} and 94K surgical datasets from our own transsphenoidal pituitary neurosurgery images. 



\noindent  \textbf{Function-calling Tuning Dataset}.
We construct 40K surgical and 140K natural function-calling conversations. The conversation is constructed with \textit{Thinking}, \textit{Calling} and \textit{Replying} to enhance the surgical function calling of VS-Assistant, specifically, learning what, how, and when to use functions. Moreover, we also construct 1K negative conversations to query non-existent objects within images and acknowledge the absence of such objects, which mitigates the hallucination of MLLM. We also construct 1K unseen conversations for test.

\noindent \textbf{Implementation Details.}
We perform the experiments in PyTorch on the NVIDIA A800 GPU. In our VS-Assistant, we select the Vicuna-7B \cite{vicuna} as the LLM base and perform the instruction tuning to produce the surgical LLM. For the visual modality, we adopt a pre-trained ViT in CLIP \cite{clip} as the visual encoder to extract visual information. In the MOP, we set the number of projectors $N$ as 8, and select $K=2$ projectors in each forward of the dynamic routing. The standard deviation $\sigma$ is set as $1$. To perform fair comparisons, we fine-tune the MLLM models on the same function-calling tuning dataset.

\noindent \textbf{Evaluation Metrics.}
To evaluate surgical function-calling performance on unseen multimodal samples, we first calculate the successful function-calling rate as queried, denoted as \textit{SR}. We further consider three types of metrics to compare with MLLM. We first the ratio of surgical triplet \cite{nwoye2023cholectriplet2021} keywords that appear in the \textit{Replying} as keyword hit rate for surgical scene analysis, denoted as \textit{KeyHit}. We also count the ratio that MLLM rejects non-existent object queries, denoted as \textit{Rej}. We adopt the BLEU@4 to evaluate the textual quality of \textit{Replying} presented to surgeons. Moreover, we further compare the detection and segmentation performance of VS-Assistant with the mean Intersection over Union (mIoU). 
\subsection{Experimental Results}
We evaluate the proposed VS-Assistant from two perspectives, including the surgical function-calling and the function performance.

\begin{table}[t]
    \begin{center}
            \caption{Comparison with state-of-the-art MLLMs on surgical function-calling.}
        \begin{tabular}{c| p{1.5cm}<{\centering} p{1.5cm}<{\centering} p{1.5cm}<{\centering} p{1.5cm}<{\centering}} \hline 
            Method & SR$\uparrow$ & KeyHit$\uparrow$ & Rej$\uparrow$ & BLEU@4$\uparrow$\\
            \hline
            Qwen-VL \cite{qwen} &$84.22$&$24.04$&$4.17$&$23.17$\\
            MiniGPT-v2 \cite{minigpt4v2} & $91.28$& $51.92$&$37.50$&$25.55$\\
            mPLUG-Owl 2 \cite{owl} & $75.12$&$29.81$&$66.67$&$22.86$\\
            LLaVA \cite{llava} & $82.19$&$39.52$&$41.67$&$28.40$\\
            LLaVA-Plus \cite{llava-plus} &$86.47$&$37.21$&$39.58$& $23.85$\\
            \rowcolor[rgb]{ .949,  .949,  .949}
            VS-Assistant & {\bf 97.36}&$\textbf{68.60}$&$\textbf{81.25}$&$\textbf{32.88}$\\ \hline
        \end{tabular}
        \label{mllm compare}
    \end{center}
\end{table}


\noindent \textbf{Comparison on Function-calling}. 
For fair comparisons, we finetune advanced MLLMs \cite{qwen,minigpt4,owl,llava,llava-plus} with the same surgical function-calling samples to enable them to accomplish the task. In Table \ref{mllm compare}, we evaluate the surgical function-calling and the output of MLLMs, and our VS-Assistant with tailored modules achieves the best performance. In particular, our VS-Assistant achieves the 97.36\% SR and 68.60\% KeyHit, revealing the advantage in surgical function-calling and surgical scene understanding, respectively. The \textit{Replying} of VS-Assistant with the BLEU@4 of $32.88$\% confirms the answer quality in language aspect, and the 81.25\% Rej reveals the advantage to reject non-existent object queries. Qualitative results of VS-Assistant are shown in Fig. \ref{model_conversation} and supplementary.


\noindent \textbf{Comparison on Function Performance}. 
For some MLLMs \cite{qwen,minigpt4v2} with the capability of object detection, we fine-tune them with surgical scene data, but this leads to limited performance. Benefiting from function-calling capability, VS-Assistant uses advanced surgical functions \cite{wang2023yolov7} and achieves superior mIoU performance in both detection and segmentation of surgical instruments, which is scalable to incorporate upgraded surgical functions in the future.


\begin{table}[t]
\begin{minipage}{0.43\linewidth}
\centering
%
%

\caption{Comparison of mIoU on detection and segmentation.}
\scalebox{0.78}{
\begin{tabular}
{c|  p{1.2cm}<{\centering} p{1.2cm}<{\centering} } \hline 
            Method &Det.$\uparrow$ & Seg.$\uparrow$  \\
            \hline
            Qwen-VL \cite{qwen}&59.68&-\\
            MiniGPT-v2 \cite{minigpt4v2}&49.44&-\\
            TransUNet \cite{transunet} &-& 90.97\\
            nnU-Net \cite{nnunet} &-&91.86 \\
            \rowcolor[rgb]{ .949,  .949,  .949} VS-Assistant&$\textbf{92.01}$&$\textbf{92.59}$\\ \hline
    \end{tabular}
    }
    \label{base detect}
\end{minipage}
\hspace{2.5em}
\begin{minipage}{0.43\linewidth}
\centering
    
    \caption{Ablation study on surgical LLM and function-calling tuning.}
    \scalebox{0.78}{
\begin{tabular}
{p{0.9cm}<{\centering} p{1cm}<{\centering}  p{1.5cm}<{\centering} p{1.2cm}<{\centering}|c c} \hline 
            \multicolumn{2}{c}{LLM} &\multicolumn{2}{c|}{FC Tuning} & \multirow{2}{*}{SR$\uparrow$} & \multirow{2}{*}{KeyHit$\uparrow$}\\
            Vanilla&\text{Tuned}& \textit{w/o} CoT & \textit{w/} CoT & \\
            \hline
            
            \checkmark& & \checkmark&& 89.47&61.69\\
            
            \checkmark& &&\checkmark& 94.73&64.52\\
            
            &\checkmark &\checkmark&& 92.10&65.10\\
            
            &\checkmark &&\checkmark&$\textbf{97.36}$&\textbf{68.60}\\
            
            \hline
  \end{tabular}
  }
  \label{base_ab}
\end{minipage}
\end{table}

\subsection{Ablation Study}
We first investigate the impact of surgical LLM tuning and the CoT in function-calling (FC) tuning in Table~\ref{base_ab}, which confirms both surgical knowledge and CoT conversations are crucial for function-calling. Then, we study the impact of projector number in MOP and find that a suitable dynamic routing is helpful for the task, as shown in Table~\ref{mop_ab}. We further study the scalability of function numbers in Table~\ref{tool_number_ab}, and confirm that the VS-Assistant can accurately call a certain number of surgical functions to meet the needs of surgeons.


\begin{table}[t]
\begin{minipage}{0.43\linewidth}
\centering
\caption{Ablation Study of MOP.}
\scalebox{0.8}{
\begin{tabular}
{c  | p{0.8cm}<{\centering} p{0.8cm}<{\centering} p{0.8cm}<{\centering}  p{0.8cm}<{\centering} p{0.8cm}<{\centering}} \hline 
            \multirow{1}{*}{Projector Num.}
            &1&2&4&8&16 \\\hline
            SR$\uparrow$ & 95.57&96.12& 95.68 & {\bf 97.36}& 77.68\\
            KeyHit$\uparrow$ & 58.47& 63.26 &64.82 & {\bf 68.60} &46.32\\ \hline
        \end{tabular}
        }
        \label{mop_ab}
\end{minipage}
\hspace{2.5em}
\begin{minipage}{0.43\linewidth}
\centering
    \caption{Function scalability.} \scalebox{0.8}{
\begin{tabular}
{c  | p{0.9cm}<{\centering} p{0.9cm}<{\centering} p{0.9cm}<{\centering}  p{0.9cm}<{\centering} p{0.9cm}<{\centering}} \hline 
        Func. Num. 
            &2&3&4&5&6 \\\hline
            SR$\uparrow$&98.68&97.36&93.42&91.96&90.26\\
            KeyHit$\uparrow$&68.92&68.60&66.51&63.22&61.86\\
            \hline
  \end{tabular}
  }\label{tool_number_ab}
\end{minipage}
\vspace{-5pt}
\end{table}



                



\section{Conclusion}
To understand the surgeon's intentions and support the surgical process, we propose a VS-Assistant to complete a series of surgical tasks on the demand of surgeons. Specifically, to achieve superior surgical multimodal understanding, we devise a MOP module to align the surgical MLLM in VS-Assistant to balance the natural and surgical knowledge. Moreover, we devise a surgical Function-Calling Tuning strategy to enable the VS-Assistant to understand surgical intentions, and thus make a series of required surgical function calls to meet the needs of the surgeons. Extensive experiments on neurosurgery data confirm that our VS-Assistant can accurately understand the surgeon's intention and achieve overwhelming performance in textual analysis and visual tasks.

%
%
%
\bibliographystyle{splncs04}
\bibliography{mybibliography}
%




\end{document}